\documentclass[journal]{IEEEtran}

\usepackage{lscape}
\usepackage{tabularx}
\usepackage{amsmath}
\usepackage{amssymb}
\usepackage{hyperref}
\usepackage{gensymb}
\usepackage{subfig}
\usepackage[dvipsnames,svgnames,x11names]{xcolor}
\usepackage{caption}
\usepackage{cite}
\usepackage{makecell}
\usepackage[markup=default,commandnameprefix=always]{changes}

\definechangesauthor[name={Ahmed Abdulkadir}, color=CadetBlue]{AA}

\newcommand{\DD}{\mathcal{D}}
\newcommand{\TT}{\mathcal{T}}

\newcommand{\XAI}{XAI}

\usepackage{enumitem}
\setlist[itemize]{label=-}

\title{Explainable, Domain-Adaptive, and Federated Artificial Intelligence in Medicine} \author{Ahmad~Chaddad*, Qizong lu, Jiali Li, Yousef Katib, Reem Kateb, Camel  Tanougast, Ahmed Bouridane, Ahmed Abdulkadir
\IEEEcompsocitemizethanks{\IEEEcompsocthanksitem A. Chaddad, Q. Lu, J. Li are with the School of Artificial intelligence, Guilin Universiy of Electronic Technology, China. A. Chaddad is with The Laboratory for Imagery, Vision and Artificial Intelligence, Ecole de Technologie Superieure, Montreal, Canada.\\
Y. Katib is with the college of medicine, Taibah University, Madinah, Saudi Arabia.\\
R. Kateb is with college of computer science and engineering, Taibah University, Madinah, Saudi Arabia.\\
C. Tanougast is with the Laboratory of Design, Optimization and Modeling Systems, University of Lorraine, Metz, France.\\
A. Bouridane is with the Center for Data Analytics and Cybersecurity (CDAC), University of Sharjah, United Arab Emirates.\\
A. Abdulakadir is with the Laboratoire de recherche en neuroimagerie, Centre Hospitalier Universitaire Vaudois, Lausanne, Switzerland.

\protect
*Corresponding author: Ahmad Chaddad, 

Email: ahmad8chaddad@gmail.com, ahmadchaddad@guet.edu.cn}
\thanks{Manuscript received November , 2021; revised August -, 2022.}}

\begin{document}

\maketitle

\begin{abstract}
  Artificial intelligence (AI) continues to transform data analysis in many domains. Progress in each domain is driven by a growing body of annotated data, increased computational resources, and technological innovations. In medicine, the sensitivity of the data, the complexity of the tasks, the potentially high stakes, and a requirement of accountability give rise to a particular set of challenges. In this review, we focus on three key methodological approaches that address some of the particular challenges in AI-driven medical decision making. (1) Explainable AI aims to produce a human-interpretable justification for each output. Such models increase confidence if the results appear plausible and match the clinicians expectations. However, the absence of a plausible explanation does not imply an inaccurate model. Especially in highly non-linear, complex models that are tuned to maximize accuracy, such interpretable representations only reflect a small portion of the justification. (2) Domain adaptation and transfer learning enable AI models to be trained and applied across multiple domains. For example, a classification task based on images acquired on different acquisition hardware. (3) Federated learning enables learning large-scale models without exposing sensitive personal health information. Unlike centralized AI learning, where the centralized learning machine has access to the entire training data, the federated learning process iteratively updates models across multiple sites by exchanging only parameter updates, not personal health data. This narrative review covers the basic concepts, highlights relevant corner-stone and state-of-the-art research in the field, and discusses perspectives.
  \end{abstract}


\IEEEpeerreviewmaketitle

\section{Introduction}
Artificial intelligence (AI)--understood here as the capability of computers to transform input data to elicit an appropriate response \cite{Shapiro.1992}, has a huge potential to transform many aspects of our lives.
The methodological advances that drive the progress generally focus on core issues like solving specific tasks more accurately\cite{ImageNet}, creating models that generalize to unseen data \cite{9023664}, or addressing fundamentally new tasks\cite{AlphaStar}.
Within these innovation cycles, the application of medical data faces particular challenges. 
Personal health related information is sensitive and therefore not readily sharable. When it is shared, the data itself is highly complex and heterogeneous. There is not only a large variation between individuals, but there is also measurement noise and systematic effects due to acquisition instruments and protocols. This noise, uncertainty, and heterogeneity in the data and labels creates distinctly challenging sets of problems. The goal of AI in medicine is to support the clinical decision process. The decisions in clinical processes can have grave consequences, sometimes observed with a substantial delay and usually uncorrectable. Furthermore, state-of-the-art AI models tend to have a large number of parameters, which are necessary to learn complex interaction and generalize well. Thus, even when being deterministic, AI models do not behave intuitively and may
fail or behave unexpectedly \cite{Nguyen.2015}. This perception of the \textsl{black-box behavior} leads to subjective mistrust and objectively limited reliability of an unknown extent.

%Explainable Artificial Intelligence (XAI):
With the growing availability of extensive medical data and the increase in computational power, AI has become more important in clinical practice for early disease detection, accurate diagnosis, predictions, and prognosis \cite{varoquaux2022machine}. However, AI models--mostly instantiated as deep artificial neural networks, are sufficiently complicated to be non-intuitive. Thus, the computer models are perceived as a \textsl{black boxes} \cite{25}. Then, the black-box nature leads to mistrust of predictive models by clinicians \cite{20}. For this reason, explainable artificial intelligence (explainable AI) was proposed in order to make AI models more transparent and easier to understand and interpret \cite{21}.
Explainable AI can thus complement human perception of the naked data\cite{holzinger2021next,holzinger2022information} and will contribute to making AI models in medicine more trusted.

%Domain Adaptation(DA):
Computer-aided diagnosis is increasingly involved in clinical decision processes to advance the goal of personalized medicine \cite{13}. However, some challenges limit its application in clinical practice, such as domain transfer \cite{guan2021domain}. Specifically, it assumes that the training and test set are independent and identically distributed (IID) using machine learning (ML) models. In reality, the distributions of the training and test set may not be identical. This leads to domain shift when we transfer the model \mbox{\cite{15}}. Despite advanced deep learning models \cite{16}, the error increases proportionately with different distributions between the training data set and the test data set \cite{guan2021domain}. Therefore, how to solve the domain transfer problem is a key to applying AI to clinical tasks. Domain adaptation (DA) breaks the traditional assumption that the distribution of the training and test data set must be consistent with ML \cite{17}. The data set is divided into source and target data in this context. The source data is related to the task to be solved, and the target data (with no labels or only a few labels) is related to the task. Note that there are always domain shifts between two domains \cite{17}.
In medical imaging, there are significant differences within data sets of the same type/modality due to various equipment, sites, protocols, etc \cite{18}. Domain adaptation techniques aim to reduce these differences. 

More and more attention has been paid to data privacy in recent years \cite{46}. In clinical topics, personalized models require protection of personal health information \cite{45}. Federated learning (FL) can integrate data from private sources without sharing private data itself \mbox{\cite{44}}. Specifically, FL protects users' privacy primarily by exchanging non-personal data such as updated models between agents (clients) of the learning federation \cite{43}. 

\begin{figure}[ht]
  %\centering 
  %\includegraphics[width=0.49\textwidth]{F00.pdf}
  \includegraphics[width=0.45\textwidth]{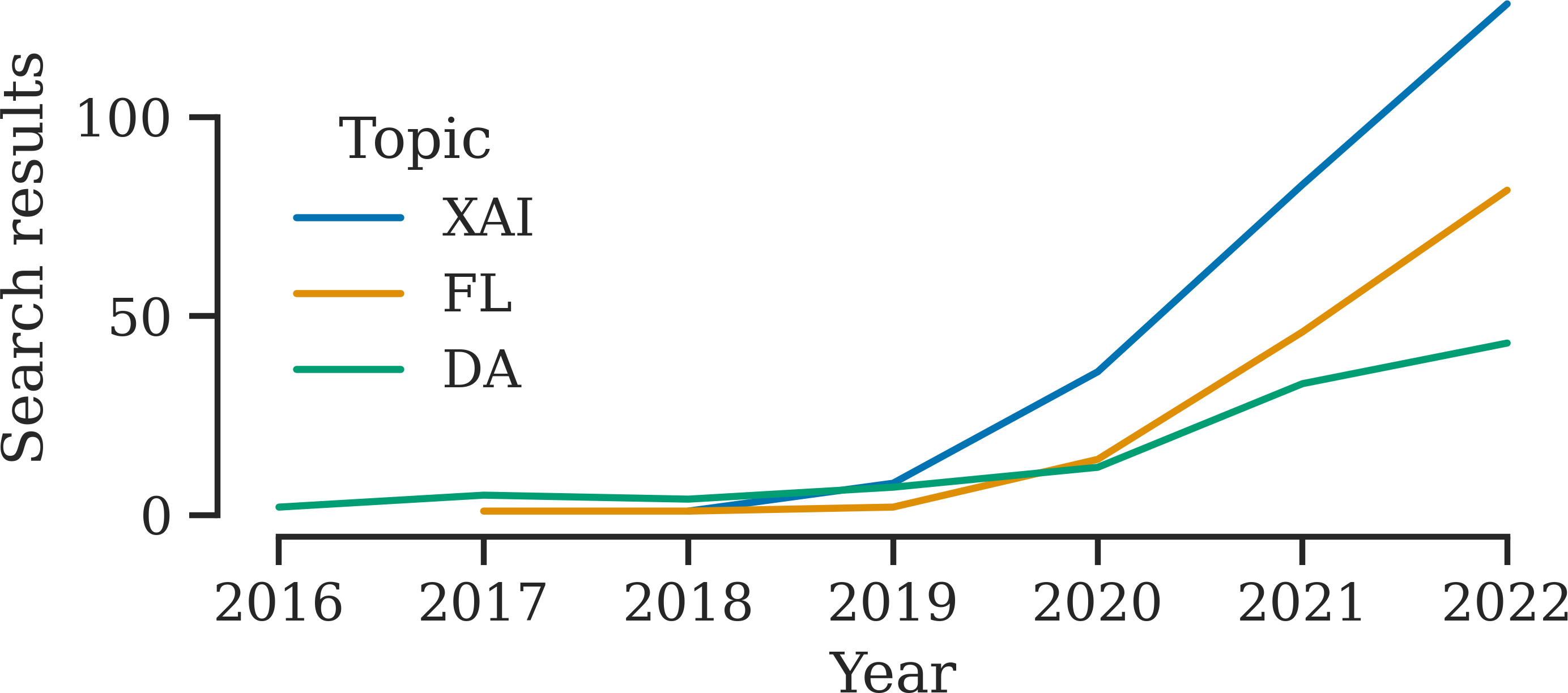}
\caption{Number of publications indexed on PubMed (https://pubmed.ncbi.nlm.nih.gov)
that matched the search queries related to the topics in this survey.\\
FL: \texttt{“federated learning” AND ((medicine) OR (healthcare))}, 
XAI: \texttt{((“explainable AI”) OR (“explainable artificial intelligence”)) AND ((medicine) OR (healthcare))}, 
DA: \texttt{(“domain adaptation”) AND ((medicine) OR (healthcare))}. The number for the year 2022 is a projection based
on the assumption that publications appear at the same rate for the
remaining two months of the year 2022.
}

\label{fig0}
\end{figure}

This paper aims to provide a brief narrative review of explainable/interpretable, domain-adaptive, and federated AI for clinical applications. Fig. \ref{fig0} illustrates a yearly growing number of publications in the field.
These search results indicate a clear trend.
However, it is worth noting that the search technique is not
100 percent sensitive because synonymous terms may be used for similar techniques; for instance, transfer learning is a type of DA. In addition, some research works may discuss one of the topics superficially yet being counted.

This review provides a brief overview of \XAI, DA and FL and
categorizes techniques within each field.
For each topic, we motivate its use in medical research and healthcare applications and discuss challenges and potentials for feature research.

The remainder of this paper is structured as follows.
Section \ref{S2} consists of the fundamental concepts for the three methodological approaches.
We discuss in Section \ref{S3} the application of \XAI, DA, and FL to medical
research and healthcare.
Finally, Section \ref{S4} concludes by providing perspectives of challenges and untapped potential.

\section{Methodological foundation and background} \label{S2}
\subsection{Explainable artificial intelligence}
Human-centric medical decision processes are designed to be trustworthy
because of the high stakes that are involved.
Isolated outputs from AI models derived from deterministic, yet highly non-linear networks that consist of artificial neurons organized in many--up to a thousand and more, layers \cite{96} leave no room for interpretation and provide no explanation of the prediction.
AI models are therefore casually considered as black boxes because the “How?” and “Why?” a machine arrives at its conclusion is in general not understandable by humans \cite{Yang.2022}.
The field has a rich and nuanced taxonomy that is not always used consistently.
Explainability and interpretability--sometimes used interchangeably,
relate to human-focused understanding
\cite{104,Murdoch.2019,BarredoArrieta.2020}.
A machine is understandable if their decisions are based on descriptive and
relevant information that is recognizable by humans.
In this context, \textsl{descriptive} means how well the interpretation method
captures the learned relationships and \textsl{relevant}
means how useful the interpretation is\cite{Holzinger.2021}.
Explainable AI models strive for high causability--the extent to which a human can casually
understand the decision making process of artificial intelligence\cite{9548137}.
Explainability and related research fields are elements of building
reliable and trusted AI systems \cite{holzinger2021next}.

%There is no unambiguous definition of \emph{explainability}, however, which sometimes is also referred to as \textsl{interpretability} and/or \textsl{understandability}. It leads to a new term issue: How can we use these words properly? For example, a clear division criterion has been discussed between 'interpretability' and 'explainability' and conditions have been defined for each of these terms \cite{BarredoArrieta.2020}. Moreover, explainability represents  completeness, transparency, and causality \cite{9548137}. More definitions of explainability have been considered in \cite{104}, which provides a standard use of the term $explainability$. The main question in this context is how we can build an explainable model. 

XAI elements are injected either into the model architecture or are external.
The former leads to models that intrinsically increase transparency of the model,
whereas the latter introduces explainability post-hoc \cite{105}.
Transparency is achieved by highlighting relevance of features, simplification, and visual data
that is relevant and meaningful for the practitioners\cite{mohanty2022comprehensive}.

\subsection{Domain-adaptive artificial intelligence}
\begin{figure}[ht]
  \centering 
  \includegraphics[width=0.4\textwidth]{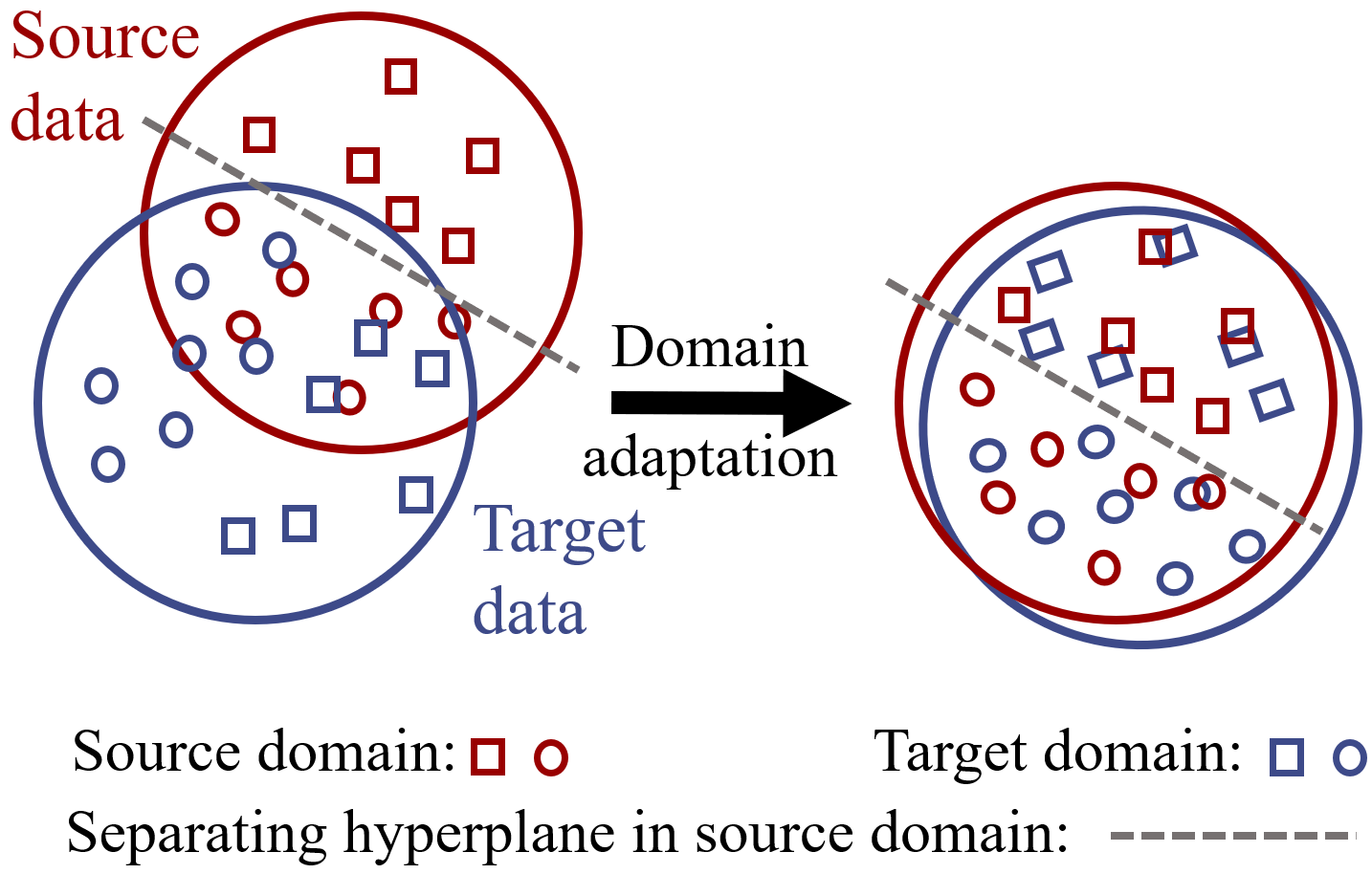}
\caption{Illustration of homogeneous domain adaptation in which the source and target distributions consist of the same features but have different distributions.
Red and blue colors represent the source and target data, respectively.
Circles and squares represent class labels.
Domain adaptation aims to applying models (dashed line) to data from a target domain (blue) that were learned using data from a source domain (red).
In the example shown, the domain adaptation consists of shifting and rotating the data points such that after the transformation, the source, and target domains overlap.
The separating classification hyperplane, shown as grey dashed line,
only accurately classifies the data points from the target data after domain adaptation.}
\label{figDA}
\end{figure}

Generalizability is the implicit ability to use a trained predictive model from one or more source domains in a different, but related, target domain.
In contrast, domain adaptation is achieved through explicit modelling of domain differences.
Assume we train a classifier model using data from a source domain $\DD_s$ for some task $\TT_s$, and to learn the same or similar task $\TT_t$ using target data $\DD_t$.
If the two domains have the same distribution ($\DD_s \sim \DD_t$) and the tasks
are identical, then the model trained for task $\TT_s$ on $\DD_s$ can also be used for the task $\TT_t$ in $\DD_t$.
If the two domains are not from the same distribution ($\DD_s \not\sim \DD_t$),
then the models trained on the source domain $\DD_s$ are likely to perform worse on $\DD_t$.
DA is utilized to exploit similarities of tasks and learn differences in data
domains to produce models for task $\TT_t$ in $\DD_t$ \cite{csurka2017domain}.
Figure \ref{figDA} shows an example of homogeneous domain adaptation, this is
when the same task is applied to the same features that have a different distribution.
Other types of DA methods are outlined in the taxonomy below.

%%%%%%%% AHMAD not finished it yet

\paragraph{Supervised, semi-supervised, and unsupervised domain adaptation} Depending on the availability of annotation, DA applications can be classified as supervised \cite{koniusz2018museum}, semisupervised \cite{34}, and unsupervised \cite{212}.
In supervised DA models, labels of all training instances are available. On the other hand, in semi-supervised learning, only a part of the training data is annotated. Learning with no labels is considered unsupervised. Specifically, most studies focus on unsupervised DA models due to insufficient numbers of labeled samples. For example, a novel unsupervised DA algorithm proposes to eliminate the domain differences for reducing the label noise \cite{27}.
While,  a deep domain adaptation method is proposed to transfer domain knowledge from a well-labeled source domain to a partially labeled target domain \cite{26}.
This method minimizes the differences in the domain that have an important clinical impact for disease diagnosis.

\paragraph{Single- and cross-modality} Medical image derived from many types of scanner (e.g., magnetic resonance imaging, CT, X-ray) that are related to texture differences between the source and target domains \cite{kim2020learning}. This phenomenon is called the modality difference. When based on acquisition, DA methods are categorized as either single- or cross-modality DA. In single-modality DA, the adaptations happen within a single type of data\cite{yan2019domain}.
In clinical practice, medical images are acquired on hardware from different vendors and at different centers \cite{argelaguet2020mofa+}.
Because different modalities provide complementary information and vary in acquisition
cost and expertise, a diagnosis can be established with different modalities.
Cross-modal DA helps to make models applicable across modalities.
For example, some models can be transferred between CT and MRI images\cite{29}.
In addition, bidirectional cross-modality DA between MRI and CT images has been applied using the synergistic image and feature alignment framework (SIFA) \cite{28}.
Despite the practicability of DA, there are always remaining risks of inherent biases
in the data\cite{Larrazabal.2020} or unfairness in models\cite{Du.2021}.

\paragraph{Single- and multi-source domain adaptation}Depending on the number of source domains, it is possible to classify DA into single-source DA (SDA), multi-source DA (MDA) and source-free DA \cite{186}. For example, trained models using single-source domains lack generalization.
Multi-source DA collects labeled data from multiple sources with different distributions \cite{32} and implicitly learn to be robust. For example, MDA learning can alleviate the difficulties caused by aligning multimodal characteristics between multiple domains \cite{31}.
Although MDA is advanced, MDA applications are challenging due to the heterogeneity of data between domains. For this reason, most existing studies are based on SDA \cite{30}. Another issue could be related to the source domains, cause to data privacy or lack of storage space for small devices \cite{48}. A generalized source-free domain adaptation (G-SFDA) was proposed to possess source domain features via transfer learning \cite{51}.

\paragraph{One- and multi- step DA} On the basis of the distance between the source and the target domain, domain adaptation methods can be roughly categorized into two types: one-step and multistep domain adaptation methods. The one-step DA methods often simply align the feature distributions of the two domains by minimizing their distances. But it is difficult for the network to minimize the huge domain discrepancy. Specifically, in remote domain transfer learning, when the distance between the source domain and the target domain is too long \cite{56}, it is impossible to achieve and can even affect the performance of the learning model in the target domain. This situation is called negative transfer \cite{185}. In order to avoid this situation, it is necessary to find an “intermediate bridge” to connect the source and the target domain. That is called multistep DA. It can be transformed into one-step domain adaptation by choosing an appropriate intermediate domain. For example, through efficient instance selection, transfer features are successfully applied between remote domains and effectively solve the remote domain problem \cite{53}.

%v2
\paragraph{Shallow and deep DA} According to the type of model, we can divide the DA method into shallow and deep DA. Usually, shallow DA leverages feature engineering and is used with traditional machine learning.
%The models adapt either features, samples, or model parameters\cite{36}\textcolor{red}{unclear what this means and the reference is about DL}.
But few timely reviews the emerging deep learning based methods. Unlike the shallow DA, deep DA methods can leverage deep networks to learn more transferable representations by embedding DA in a deep learning pipeline \cite{wang2018deep}.
In addition, deep DA integrates a feature into the learning model to improve performance metrics \cite{guan2021domain}.

These methods may reduce the differences in the domain that can have
a positive impact in clinical research.
Specifically, more effort is required to fully apply such methods in clinical topics to avoid bias.

%V2
\subsection{Federated learning} Federated Learning (FL) is a way of distributed collaborative machine learning in which the training data is not shared. This concept of data-private collaborative learning \cite{Sheller.2020}, that initially was motivated in the context of mobile devices in which each one has only a small fraction of the data, equally applies to medical application in which the data is distributed across hospitals \cite{Sarma.2021}.

This framework works by collaborating multiple devices to update a machine learning model without sharing their private data under the supervision of a central server. An example in medicine is the COVID-19 chest scan engine, which was built by many researchers and medical practitioners from different parts of the world. In the COVID-19 pandemic, along with increased privacy concerns, FL is mainly useful to support the diagnosis and detection of COVID-19. This was done by coordinating massive hospitals to build a common AI model \cite{kumar2021blockchain}.

We now give a general formal definition of FL.
Let $N$ be the number of data clients--for example hospitals,
$\left\{\mathcal{H}_1,\mathcal{H}_2, \ldots, \mathcal{H}_N\right\}$, all of whom wish to contribute to the training of a predictive model by combining their respective data $\left\{\mathcal{D}_1,\mathcal{D}_2, \ldots, \mathcal{D}_N\right\}$. A traditional method is to combine all data and use $\mathcal{D}=\mathcal{D}_1 \cup \mathcal{D}_2, \ldots \cup \mathcal{D}_N$ to train a model $\mathcal{M}_{all}$. A FL is a learning process in which the data clients collaboratively train a model $\mathcal{M}_{FL}$, in which process any data client $\mathcal{H}_i$ does not share its data $\mathcal{D}_i$ to others. In addition, the accuracy of $\mathcal{M}_{FL}$, denoted as $\mathcal{V}_{FL}$ should approach the performance of $\mathcal{M}_{all}$, $\mathcal{V}_{all}$. Formally, let $\delta$ be a non-negative real number, if
\begin{equation}
\left|\mathcal{V}_{Fl}-\mathcal{V}_{all}\right|<\delta
\end{equation}
Then, the federated learning algorithm has an accuracy loss of $\delta$.

FL is particularly attractive for smart healthcare, by coordinating multiple hospitals to perform an AI training without sharing raw data \cite{crowson2022systematic}.
However, FL is an emerging research area and has not yet gained much trust in the medical community; therefore, it has required more research, particularly with respect to security and privacy aspects.
Although this framework sounds ideal in theory, it is not immune to attacks. Moreover, its current development is not mature enough to be expected to solve all privacy issues by default. The most specific security threats facing the FL framework are communication bottlenecks, poisoning, backdoor attacks, and inference-based attacks, which can be considered the most critical to FL privacy \cite{mothukuri2021survey}. In this context, FL is represented in three main groups 1) privacy protection \cite{mcgraw2021privacy}, 2) Effective communication \cite{wu2022communication} and data heterogeneity \cite{qu2022rethinking}. 
%Regardless of the advanced FL models, more efforts are needed to consider these models in healthcare field.

%We introduce how FL is applied for healthcare as follows.
\paragraph{Privacy protection} For generalization of the predictive model in medical data analysis, recent work focuses on considering multisite samples with multidimensionality. However, data sharing between institutions has been controlled by laws such as the United States Health Insurance Circulation and Accountability Act (HIPAA) \cite{42} and the European General Data Protection Regulation (GDPR) \cite{41}). FL demonstrates its feasibility in sharing and protecting the privacy of 20 medical institutions and centers \cite{65}. FL is represented by models that are trained simultaneously at each site. These models are periodically aggregated and redistributed, requiring only the weight transfer of learned model between institutions. This step eliminates the requirement of sharing data directly \cite{70}. Although FL can efficiently protect the data privacy, their application still bares some security-related risks \cite{77}.

\paragraph{Effective communication} Communication is a critical bottleneck in federated networks \cite{165}. FL network can consist of thousands of agents contributing model updates. To obtain high-performance models, the multi-participant situation is inevitable. In this context, network communication is slower than local computation by many orders of magnitude due to limited resources such as bandwidth, energy, and power. To solve this problem, two key aspects can be used: 1) reducing the total number of communication rounds and 2) reducing the size of the messages transmitted in each round \cite{164}. It can perform client-side filtering, reduce the frequency of updates, and P2P (peer-to-peer) learning (e.g., compression) \cite{38}.

\paragraph{ Data heterogeneity} Currently, data samples in hospitals are not independent and distributed identically (IID). It leads to an increase in the bias of the model \cite{87}. A proposed initial shared model used for hospitals that can easily adapt it to the local dataset \cite{86}. Through intelligently choosing the participant, it can reduce the times of communication in FL and counterbalance the bias introduced by IID data. However, the data are not always IID type. To systematically and intuitively understand the current FL method in different data distribution, new studies have been conducted a comprehensive experimental comparison between the new algorithms \cite{89}. This means that FL can solve the problem of multisite data, such as in multiple clinical institutions \cite{90}.

Regardless of these advanced FL techniques, more efforts are needed to consider these models in healthcare field.

\section{Methods}\label{S3}

Deep learning models are widely applied in medicine. However, these models require more interpretability to simplify the steps of the prediction process. Despite advances in XAI models, there is a huge need to consider XAI with DA and FL when the topics are related to medicine to achieve the goal of personalized medicine. For example, DA-based XAI models have the ability to improve the performance in computer aided diagnosis \cite{walambe2022explainable}. But the same model will not be able to provide a feasible performance using multi-institutional data. For this reason, FL is a good option for data sharing under the condition of protecting data privacy \cite{57}. However, there is an additional issue related to intra- and inter- site difference of distributions. This leads to a domain shift between sites. Recent work proposes to integrate Unsupervised Domain Adaptation (UDA) into the FL framework \cite{232}. For example, UDA guides the model to learn domain-agnostic features through adversarial learning or a specific type of batch normalization \cite{56}. Furthermore, DA techniques such as mixture of experts (MoE) with adversarial domain alignment can be used to improve the accuracy of different sites in the FL learning setup \cite{58}. These techniques will be presented below in three subsections related to DA, XAI and FL.

% --------
\subsection{Current methods of explainable artificial intelligence}
\begin{figure*}[ht]
  %\centering 
  \includegraphics[width=1\textwidth]{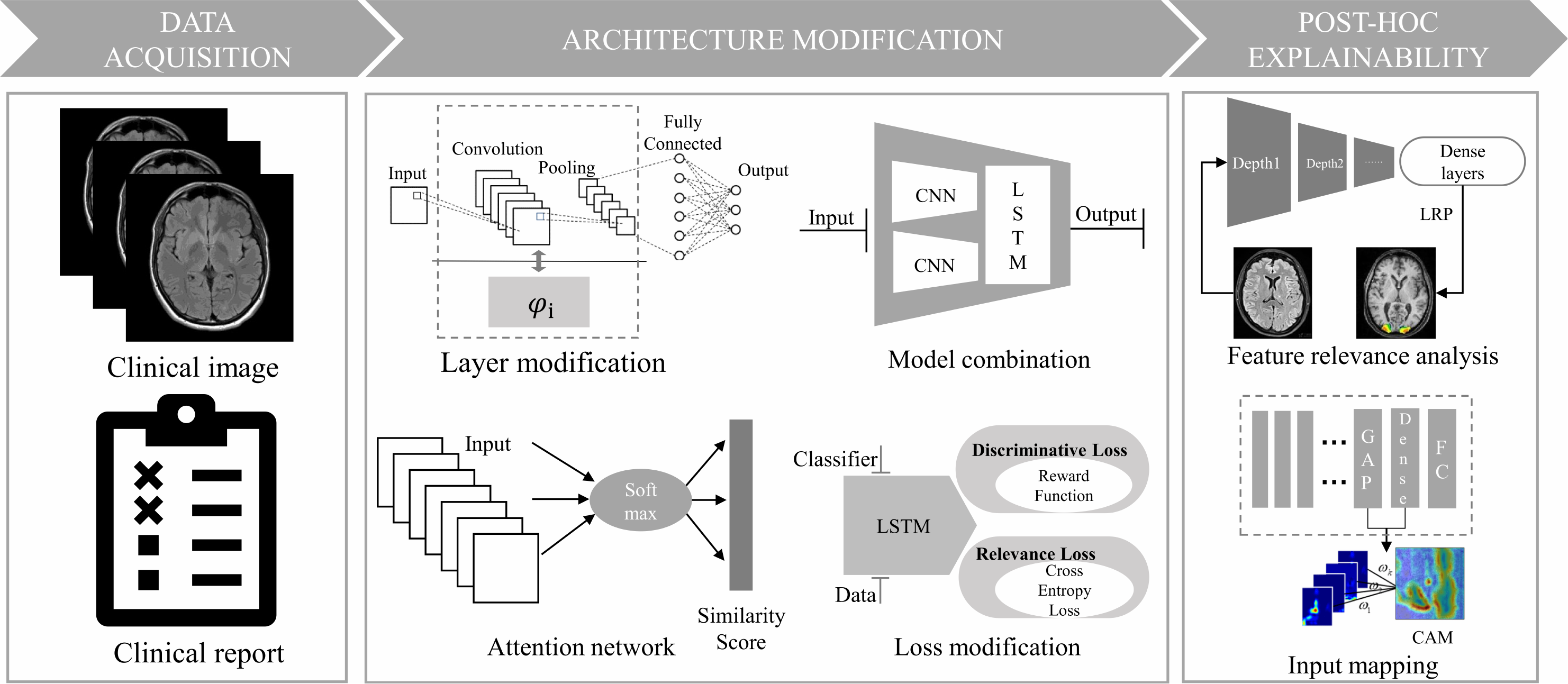}
\caption{Example of flowchart for the explainable artificial intelligence model, 1) data acquisition consists of images and clinical report for each of patients. 2) Example of four methods during modeling: it can modify layer (Layer modification) to simplify the structure or combined (Model combination) with other model to build a full explainable model. In addition, an attention network could be used for visualization and even to modify the loss function. 3) The two most used methods after modeling: feature relevance analysis and visual explanation.}
\label{fig2}
\end{figure*}

With high-performance deep learning models of increasing complexity,
the way specific results were obtained are not accessible to humans.
A number of methods were proposed to increase explainability in recent
years, 17 of which were discussed in a recent survey \cite{holzinger2022explainable}
and a selection of which is listed in Table \ref{table2} and discussed in
following paragraphs.
These techniques are mainly divided into two categories, 1) explainability in modeling and 2) explainability after modeling (see Figure \ref{fig2}).

\paragraph{Explainability in modeling}
Explainability is derived from modifications of the model architecture.
This process can be represented in simplification (and/or quantification) layers and loss function, or by adding an explainable network and visualization module. For example, CNN architectures consist of convolutions, pooling, and fully connected layers; To increase the explainability of CNN, layers could be quantified, modified, and combined with a visual model. Replacement of convolutional layers with max-pooling layers provide better visualization without losing performance metrics \cite{109}. Another scenario related to CNN is represented by adding attention network and modifying the loss function \cite{111}. Recent studies have been quantified the convolution layers by Gaussian mixture model and/or texture features for clinical classifications \cite{chaddad2019deep,chaddad2019predicting,chaddad2020modeling,chaddad2021modeling,chaddad2021deep,chaddad2022deep}. However, there are only a few studies focused on explainable hidden layers.

\paragraph{Explainability after modeling}
It requires many experiments to balance model accuracy based on post-hoc interpretability (explainability after modeling). It is grouped into the following categories:

\emph{1) Feature relevance analysis}: Measures the importance of features in the prediction model (e.g.,  Layer-wise Relevance Back Propagation (LRP)). LRP is an explainable AI model that explores the relation between output and input data by generating separate correlation maps for each individual image. LRP with explainability concept is also used to prune CNN. It finds the most relevant units (i.e., weights or filters) with their relevance scores \cite{112}. Deep-LIFT is also a method to improve the effectiveness of feature learning and improve the interpretability of the model itself. It works by comparing the activation of each neuron to its reference activation and assigns scores according to the difference between the two \cite{li2021deep,shrikumar2017learning}.

\emph{2) Visual explanation}: To further improve model visualization, heat maps, saliency maps, or class activation methods are considered. Among these functions, Class Activation Mapping (CAM) is the most widely used. In CNN, it can improve explainable capabilities by replacing the fully connected layer with the global average pooling (GAP) layer \cite{116}. %By using GAP, we can accurately find the feature corresponding \textcolor{red}{to which original image area}. 

However, CAM requires modifying the structure of the original model by retraining the model. For example, another version of CAM known as Grad-CAM that is a deep neural network \cite{113}. It uses the gradient information that is related to the convolutional layer and decision of interest. Whether it is CAM or Grad-CAM, the visualization results they generate are not visually enough. In addition, a Score-CAM model proposes to manage the dependence on gradients by different ways for obtaining linear weights \cite{114}.

\paragraph{Explainability level} The explainability scale is divided into different categories in different situations. For example, XAI is divided into five scales according to ethology in explaining animal intentionality \cite{119}. According to the transparency of the model, XAI is divided into simulatability, decomposability, and algorithmic transparency \cite{BarredoArrieta.2020}. Furthermore, XAI with deep learning models could be grouped into 1) local and 2) global explanation. For local explanation, the model considers a local approximation of how the black box works. It mainly focuses on explanation of data instances. The global explanation describes the whole mechanism of the model. Generally, a group of data instances is needed to generate one or more explanation maps.

%% ahmad here
\begin{table*}[!ht]
\caption{Summary of the XAI models used for clinical applications.}\label{table2}
\footnotesize
%\tiny \setlength{\tabcolsep}{17pt}
\begin{tabularx}{\textwidth}{lccc}
%\begin{tabular}{lcccc}
%\begin{tabular}{L{1cm}c{1.8cm}c{2cm}c{3.8cm}c{2cm}}
\hline%
Application &  XAI / Explainability level   &  Prediction model     & Reference    \\  \hline 
 Preclinical relevance assessment   &  Model combination / G     &  GNN, FNN         & \cite{131}       \\                          
 Detection and prediction for Alzheimer’s disease   &  SHAP /  L \& G                 &  RF               & \cite{132}       \\  
 Deterioration risk prediction of hepatitis &  SHAP, LIME, PDP  / L \& G       &  RF               & \cite{133}       \\ 
 Non-communicable diseases prediction  &  DeepSHAP  / L \& G              &  DNN              & \cite{134}       \\ 
% Diagnosis of atrial fibrillation &  Grad-CAM / L              &  CNN               & \cite{135}       \\ 
 Spinal posture classification  &  LIME / L                   &  SVM               & \cite{136}       \\ 
 Classification of estrogen receptor status  &  SmoothGrad / L   &  DCNN              & \cite{137}       \\ 
 Differential diagnosis of COVID-19 &  Grad-CAM /  G           &  CNN                & \cite{138}       \\  
% Coronavirus COVID-19 detection &  Grad-CAM / L             &  VGG-16             & %\cite{139}       \\ 
% Explanation of diabetic retinopathy  &  Grad-CAM / L             &  EfficientNetB7      & \cite{140}      \\ 
% COVID-19 screening  &  Grad-CAM / L              &  VGG-16, VGG-19     & \cite{141}      \\  
% Detection of lung pathologies  &  CAM / L                &  VGG-16             & \cite{142}      \\ 
 Prediction of mortality  &  Shapley additive / L      &  RNN                & \cite{143}      \\ 
% COVID-19 detection  &  t-SNE, Grad-CAM / L         &  7 Network          & \cite{144}      \\ 
 Medical image segmentation &  Combination / L \& G   &  CNN, Attention Mechanism    & \cite{145}      \\ 
 Early detection of Parkinson's disease  &  LIME / L                    &  VGG-16             & \cite{146}      \\ 
 COVID-19 diagnosis  &  Grad-CAM ++, LRP / L      &  DNN                & \cite{147}      \\ 
 Evaluation of cancer in Barrett's esophagus  &  5 XAI methods / L    & AlexNet, SqueezeNet, ResNet50, VGG16    & \cite{148}      \\ 
 Heart anomaly detection  &  SHAP and Occlusion maps / L     &  CNN, MLP        & \cite{149}      \\   
 Distinguish VTE patients  &  DeepLIFT / L             &  ANN                & \cite{150}      \\  
 Detection of ductal carcinoma in situ &  LRP / L                 &  Resnet-50 CNN       & \cite{151}      \\ 
 Lesion prediction &  Activation Maps, Saliency Maps / L    &  U-net               & \cite{152}      \\ 
 Early detection of COVID-19 &  Saliency Maps / L         &  Fuzzy-enhanced CNN  & \cite{153}      \\ 
% Severe community-acquired pneumonia  &  SHAP, Grad-CAM++ / L        &  ——                 & \cite{154}      \\ 
 EEG emotion recognition &  SmoothGrad, saliency maps / L    &  CNN                & \cite{155}      \\ 
 Cerebral hemorrhage detection and localization &  Grad-CAM / L                &  ResNet             & \cite{156}      \\ 
 Macular disease classification &  t-SNE, model simplification / L     &  CNN                & \cite{157}      \\ 
 Automated diagnosis and grading of ulcerative colitis &  Grad-CAM / L   &  CNN                & \cite{158}      \\ 
 Diagnostic for breast cancer &  Causal-TabNet / G   &  GNN                & \cite{159}      \\ 
 Prediction of depressive symptoms &  LIME / L  &  RNN                & \cite{160}      \\ 
 Early prediction of antimicrobial multidrug resistance &  SHAP / L \& G   &  LSTM               & \cite{161}      \\ 
 Predictions of clinical time series  &  CAM / Local  &  FCN                & \cite{162}      \\  \hline%             
\end{tabularx}
\footnotesize \\
L: Local; G:Global; ANN: Artificial Neural Network; GNN: Graph Neural Networks; FNN: Fuzzy Neural Network; DNN: Deep Neural Networks; RNN: Recurrent Neural Network; FCN: Fully Convolutional Network; DCNN: Deep Convolutional Neural Network; VGG-16: GG-Very-Deep-16 CNN; VGG-19: GG-Very-Deep-19 CNN; Fuzzy-enhanced CNN: Fuzzy-enhanced Convolutional Neural Network; RF: Random forest; SVM: Support Vector Machine; MLP: Muti-Layer Perceptron; LSTM: Long Short-Term Memory network; PDP: Partial Dependence Plot; t-SNE: t-distributed Stochastic Neighbor Embedding. Seven Network: SqueezeNet, Inception, ResNet, ResNeXt, Xception, ShuffleNet, DenseNet; Five XAI methods: Saliency, guided backpropagation, integrated gradients, input gradients, and DeepLIFT;
\end{table*}

Table\ref{table2} summarizes XAI in deep learning models applied to clinical tasks. We note that CNNs are the most widely used deep learning models and gradient-based explainability is the most applied principle. The level of explainability is mostly $local$.

\textbf{Agnostic model} In addition to the previous XAI methods, Shapley Additive Explanations (SHAP) and Local explainable Model-agnostic Explanation (LIME) are agnostic to the deep learning models they are applied to.
Specifically, LIME uses explainability models (such as linear models and decision trees) to approximate the target predictions from the black-box model. It detects the relationships between the output of the black-box model by
slightly perturbing the input in the XAI model. For example, LIME demonstrated feasible explanations in artificial neural networks by altering the input feature and fitting it to a linear model \cite{117}. The SHAP explains how each feature affects the predicted value (known as Shaply). It is computed based on the average of the marginal contributions of all permutations (permutations of each feature). It provides different explanations for a particular model (e.g., feature importance visualization, linear formulation). With SHAP \cite{118}, it illustrates how the design parameters affect performance using DNN features. 

% -------
\subsection{Domain adaptation with deep learning}
We present the main steps in DA model by summarizing the existing literature. We grouped the existing methods into 1) DA-related data, 2) DA-related features, and 3) building and adversarial domain adaptation. In addition, we report these models in Table \ref{table1}. We note that the problem of domain bias occurs when the target domain is different from the source domain. This bias type limits the performance of transfer deep learning model that leads to disable the generalization of model. 

\paragraph{Domain adaptation related data}
It assumes that DA can solve the problem of domain shift. However, there is a bias due to different hospital equipment. Instance Re-weighting Adaptation is a method to solve this issue when the domain bias is small. It removes samples from the source domain that mismatch the similar distribution in the target domain. For example, samples with higher correlation with target samples are assigned larger weights \cite{189}. The assigned data are then re-entered into the training model to reduce the problems arising from domain shifts \cite{68}. This idea has been developed with a task of re-weighting mechanism that considers loss function with dynamically update weights \cite{67}. Regardless of the advances algorithms and methods used in DA related data, these models are able to only solve the problem of small domain bias.

\paragraph{Domain adaptation related features}
It is based on the feature similarity between the source and the target domain. For example, the feature transformation strategy can solve the bias problem by transferring the source and target samples from the original feature space to
the new feature representation space \cite{71}. A novel feature-based transfer learning algorithm for remote domains requires only a small portion of labeled target samples from a different domain \cite{190}. However, transfer
features can only reduce the distribution variance from a single perspective, and it is a challenge to find an optimal transfer learning method for a given dataset. For this reason, the distribution variance was reduced from multiple perspectives by applying multi-feature-based transfer learning methods \cite{191}.

%v2
\paragraph{Building an adversarial domain adaptation}
Data- and feature-based DA aims to align a labeled source domain and an unlabeled target domain, which requires accessing the source data. It will raise concerns about data privacy, portability, and transmission efficiency. Therefore, a method that can adapt source-trained models towards target distributions without accessing source data will be promising. To adapt a model, samples from the source and target domains can share similar
representations to provide similar predictions \cite{194}.
These models need to be pre-trained using the source domain data and re-trained by the target domain \cite{194}, which will increase the processing time. Therefore, these types of predictive models are not preferred. Another scenario based on adversarial domain adaptation is currently one of the mainstream methods. It is known that deep models (e.g., adversarial domain adaptation) can use the transformed features for domain adaptation \cite{230}. For example, an unsupervised method based DA used for predicting glaucoma fundus \cite{192}.
Specifically, they designed a loss function that learns the source region features and possesses the original labels unchanged. An improved adversarial domain adaptation network has been considered for tumor image diagnosis with few and/or no labels \cite{193}. 

\begin{table}[ht!]
%\centering
\caption{Summary of recent domain adaptation models used for clinical topics.}\label{table1}
\setlength{\tabcolsep}{1pt}
%\renewcommand{\arraystretch}{1.2}
%\tiny
%{0.8\textwidth}
\begin{tabularx}{0.49\textwidth}{lccccc}
%\begin{tabular}{lcllll}
\hline
Application    & Method  & Cross-domain   & Topics      & Reference      \\
\hline
 %Glaucoma          &  GAN                                &  N           &  N            & %\cite{192}   \\
 Gastric epithelial tumor          &  GAN                              &  N          &  Y           & \cite{193}   \\
 Multiple sclerosis lesions                           &  CNN                              &  N          &  N           & \cite{194}   \\
 Arrhythmia heartbeat                                 &  CNN                              &  N          &  N        & \cite{195}   \\
 %ECG classification                                   &  GAN                              &  N          &  Y           & %\cite{196}   \\
 Bio-metric identification                            &  CycleGAN                         &  Y          &  S and N  & \cite{197}   \\
% Liver                                                &  CNN(U-Net)                       &  Y          &  NA          & %\cite{198}   \\
 Medical image                                        &  DDA-Net                          &  Y          &  N             & \cite{199}   \\
% COVID-19                                             &  DASC-Net                         &  N          &  NA           & %\cite{200}   \\
 Cardiac                                              &  PnP-AdaNet                       &  Y          &  N          & \cite{201}   \\
 %ECG delineation                                      &  ARAN                             &  Y          &  N         & %\cite{202}   \\
 Brain dementia identification                        &  SVM                              &  Y          &  Y          & \cite{203}   \\
 Brain dementia identification                        &  AD2A                             &  N          &  N          & \cite{204}   \\
 Autism spectrum disorder                             &  MSDA/MVSR &  N          &  Y         & \cite{205}   \\
 Cardiac                                              &  CycleGAN                         &  Y          &  N          & \cite{206}   \\
 Lung                                                 &  ADDA                             &  N          &  N         & \cite{207}   \\
 Histopathological                                    &  HisNet-SSDA                      &  Y          &  S         & \cite{208}   \\
 Breast cancer                                        &  DADA                             &  N          &  Y         & \cite{209}   \\
 Pancreatic cancer                                    &  GCN                              &  N          &  N         & \cite{210}   \\
 Atrial fibrillation                                  &  3DU-Net                          &  N          &  Y        & \cite{211}   \\
\hline
\end{tabularx}\\
%T, F: Whether it is a different domain, T: True, F: False; 
Y: Yes; N: No; S: Semi-supervised; GAN: Generative Adversarial Network; CNN: Convolutional Neural Network; CycleGAN: Cycle-Consistent Generative Adversarial Networks; DDA-Net: Dual Domain Adaptation Network; DASC-Net: Domain Adaptation based Self-Correction model; PnP-AdaNet: Plug-and-Play Adversarial Domain Adaptation; SVM:Support Vector Machine; AD2A: Attention-guided Deep Domain Adaptation; MSDA: Multi-Source Domain Adaptation; MVSR: Multi-View Sparse Representation; ADDA: Adversarial Discriminative Domain Adaptation; HisNet-SSDA: a deep transferred semi-supervised domain adaptation; DADA: Depth-Aware Domain Adaptation; GCN: Graph Convolutional Networks;3D U-Net: three dimension U-net.
\end{table}

Table \ref{table1} reports the DA methods used for clinical applications. We find that the dominant research approaches in recent years are related to adversarial networks, with a lower dominant for DA methods based on features and data. It observes that the most of these approaches based on unsupervised DA. This is due to the limitation of medical data. These contributions provide a solid foundation for future DA methods applied in medicine.

% --------
\subsection{Federated learning modeling}
FL updates the models to achieve high performance by considering co-modeling step between different data sources (or client). FL is divided into the following categories:

\begin{figure*}
    \centering
    \subfloat[]{{\includegraphics[width=7cm]{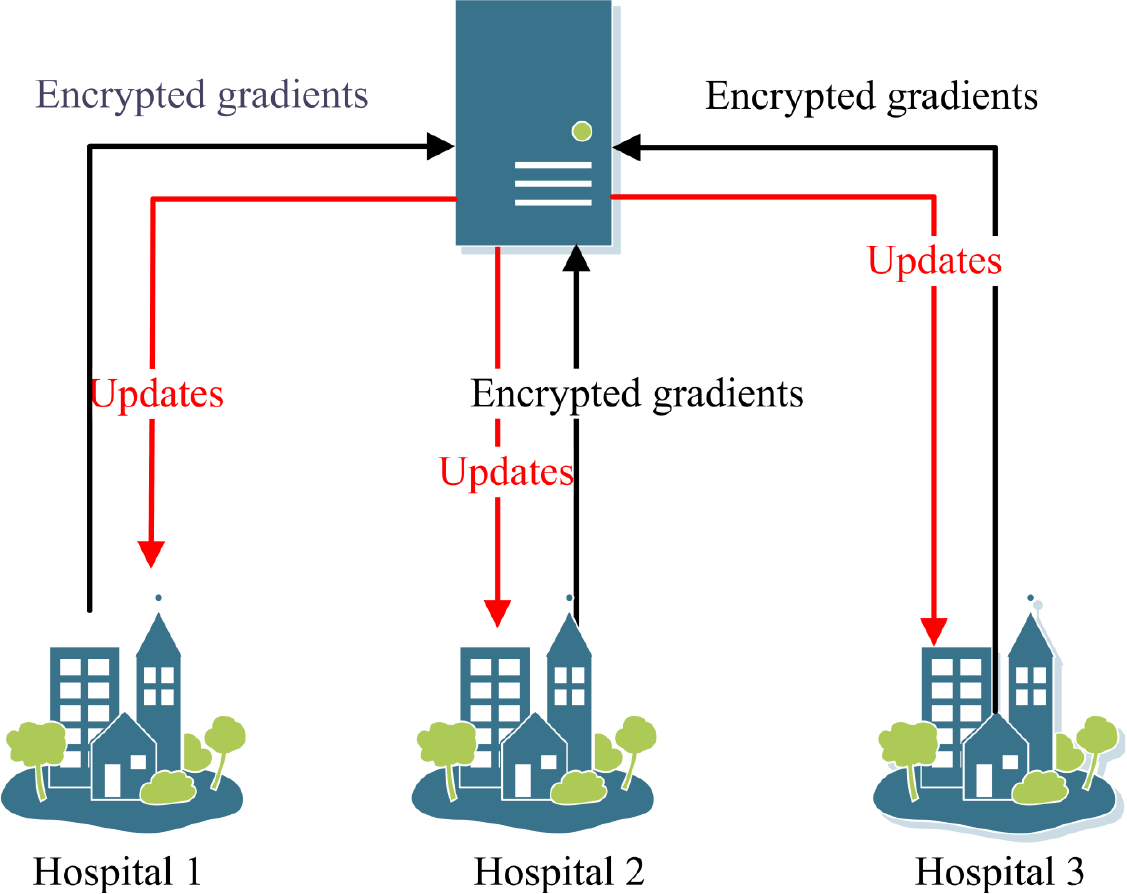} }}
    \qquad
    \subfloat[]{{\includegraphics[width=7cm]{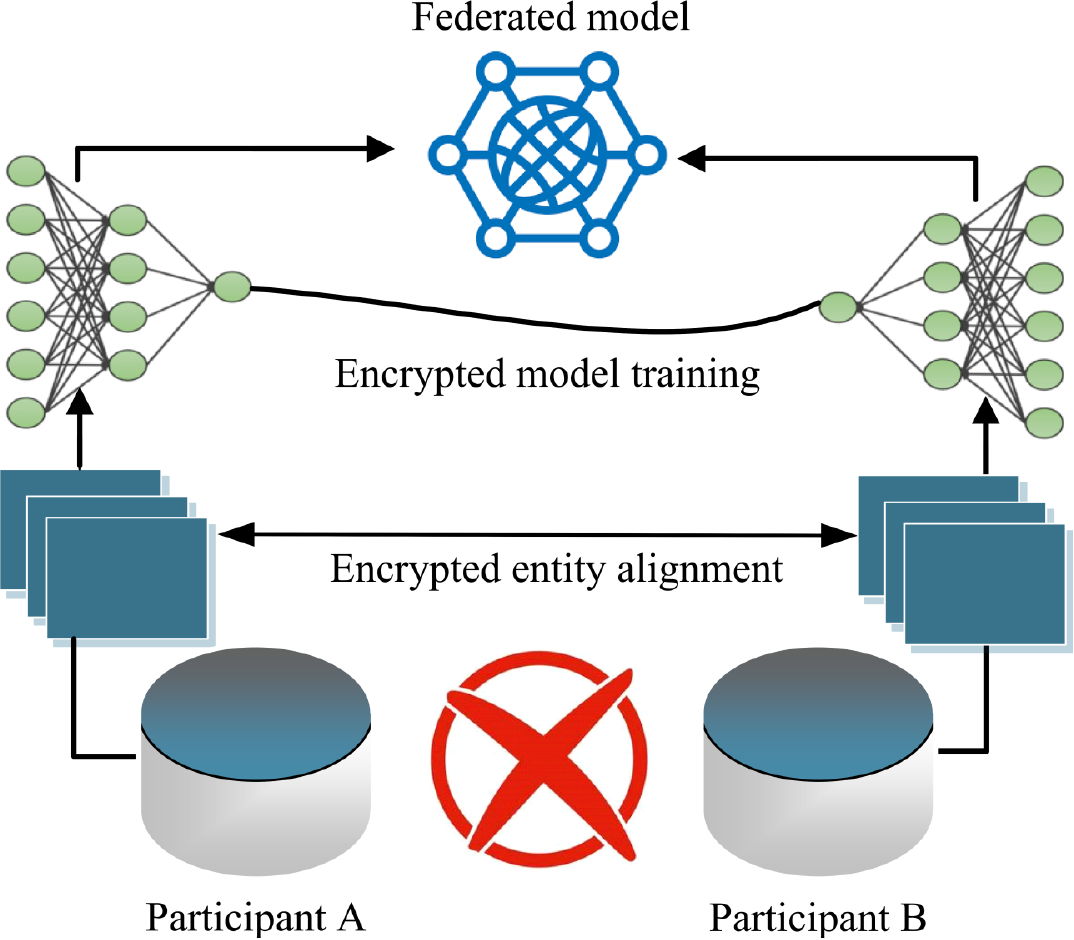} }}
    \caption{Example of federated learning based on a) horizontally and b) vertically technique; In horizontally federated learning, multiple clients (Hospitals) with the same data structure collaborate to train a predictive/classifier model with the help from server. In vertically federated learning: First, align the common users in different participants. Then, it performed the encrypted training model based on these common users.}
    \label{fig3}
\end{figure*}

%\begin{landscape}
\begin{table*}[ht!]
\caption{Summary of federate learning models used for clinical topics.}\label{table3}
%\tiny \setlength{\tabcolsep}{10pt}
\footnotesize
\begin{tabularx}{\textwidth}{lccccc}
\hline%
 Application &  Model  / Learning / optimization   &  $n$   &  Performance (Local / Central / FL models)  & Reference \\ \hline
Predicting clinical outcomes  & EXAM / HFL / ——       & 20 & AUC: 0.79/ - / 0.92       & \cite{213}   \\
Precision medicine & RETAIN / HFL /   FedAvg       & 42 & AUC: - / 0.73 / 0.62    & \cite{233}   \\
% Kappa score: 0.61 / - / 0.68    &  ——         & 7 &  HFL /   ——       & %\cite{214}   \\
Spinal cord gray matter segmentation  & DNN / HFL / FedAvg / FedPRox & 4 & Sensitivity: - / 0.76 / 0.78     & \cite{234}   \\
 Prostate cancer diagnosis  & 3D AH-Net / HFL /  ——       & 3 &  Dice: 0.812–0.872/ - / 0.89      & \cite{215}   \\
 Detection diabetic retinopathy  & AlexNet / FTL / FedAvg / FedProx & 5 & Acc: 0.81-0.92/ - / 0.65-0.90      & \cite{235}   \\
Prediction in COVID-19 patients   & LASSO, MLP / HFL /  ——       & 5 &  AUC: - / 0.719-0.822 / 0.786-0.836    & \cite{216}   \\
 Image reconstruction  & U-net / HFL  / FedAvg       & 4 &  SSIM: 0.94/ - / 0.93         & \cite{217}   \\
 Computational pathology  & CNN / HFL  /        & 3 & AUC: - / 0.985±0.004 / 0.976±0.007   & \cite{218}   \\
 Predicting outcomes in patients  & ResNet-34 / HFL / FedAvg     & 20 &  AUC: 0.79 / - / 0.92     & \cite{219}   \\
 Multimodal melanoma detection  & EfficientNet / HFL / FedAvg      & 5 & Acc: - / 0.83 / 0.83   & \cite{220}   \\
 Arrhythmia detection of Non-IID ECG   &  —— / HFL /   EWC       & 100 & F1-score: - / — / 0.80       & \cite{221}   \\
Joint training   &  ——  / HFL / FedACS        & 3 & Acc: - / — / 0.70     & \cite{222}   \\
 MultiView training  &  CNN / VFL/ ——       & 2 & Acc: 0.62/ - / 0.71      & \cite{223}  \\
 Prognostic prediction  &  ELM / HFL / ——       & 4 &  Acc: - / 0.89 / 0.8899      & \cite{224}  \\
 Designing ECG monitoring healthcare system &   CNN / FTL /  ——     & 3 & Acc: - / 0.94 / 0.92      & \cite{57}  \\
 Multi-center imaging diagnostics  &  3D-CNN, ResNet-18 / HFL / ——     & 4 & AUC: - / 0.85 / 0.86     & \cite{226}  \\
% AUC: - / 0.82-0.84 / 0.79-0.81     &  ——         & 5 &  HFL /  FedAvg     & %\cite{227}  \\
 COVID-19 detection & ResNet / HFL  / FedMoCo       & 3 & Acc: - / 0.96 / 91.56              & \cite{236}  \\ 
\hline%                                                                      
\end{tabularx}%
\\
\label{tab:multicol}

Local model: training alone in client; Central model: use all client data for training; FL model: model under FL training methods; $n$: number of institutions; EXAM: Electronic Medical Record (EMR) chest X-ray AI model; HFL: Horizontally Federate Learning; VFL: Vertically Federated Learning; FTL: Federated Transfer Learning; DNN: Deep Neural Networks; CNN: Convolutional Neural Network; 3D-CNN: three-Dimensional Convolutional Neural Network; AlexNet, EfficientNetB7, ResNet, ResNet-34, ResNet-18, U-net: a class in CNN; AUC: Area Under Curve; Acc: Accuracy; 3D AH-Net: 3D Anisotropic Hybrid Network; LASSO: Least Absolute Shrinkage and Selection Operator; MLP: Multilayer Perceptron; Dice: Dice coefficient range; Kaapa score: is a measure of classification accuracy; SSIM: Structural Similarity Index Measure; PSNR: Peak-Signal-to-Noise Ratio; ELM: Extreme Learning Machine; RETAIN: has two parallel RNN branches merged at a final logistic layer; EWC: elastic weight consolidation; FedAvg: Federated Averaging; FedACS: FedAvg with adaptive client sampling; FedMoCo: a robust federated contrastive learning framework; FedProx: A optimal method to tackle heterogeneity in federated networks.
\end{table*}
%\end{landscape}

\paragraph{Horizontal federate learning}
Horizontal federated learning (HFL) is a system in which all the parties share the same feature space. It does the training process as follows: 1) Participants first download the latest model from the server individually and train the model based on the local data, 2) The trained model with its parameters (i.e., cryptographic gradients) are uploaded to the server, and 3) The server aggregates the gradients of each user and returns the updated model to each participant to update local models \cite{60}. With this advantage, HFL is widely used in clinical topics \cite{120}.

\paragraph{Vertical federated learning}
Vertical Federated Learning (VFL) lets multiple parties that possess different attributes (e.g., features and/or labels) of the same data entity (e.g., a person) to jointly train a model. Its learning process consists of 1) encrypted sample alignment and 2) an encryption training model. 1) encrypted sample alignment: Using encryption-based user ID alignment technology to align common users. These common users data are then considered for 2) encryption training model. However, in learning process, VFL requires multiple communications between all participants that need a high processing time. For this reason, many studies have developed an asynchronous training algorithm based on vertically partitioned data \cite{125}. We note that many existing methods focus on HFL when all feature sets and labels are available.

\paragraph{Federated transfer learning}
Federated transfer learning (FTL) generalizes the concept of federated learning and emphasizes that collaborative modeling learning can be performed on any data distribution and entity \cite{127}. For example, participants first compute and encrypt their intermediate results with their gradients and losses. The gradients and losses are then collected and decrypted by a collaborator. Finally, the participants receive the aggregated results that are used to update the respective models. We note that FTL uses homomorphic encryption for security and the polynomial difference approximation for 1) privacy and 2) avoid data leakage \cite{60}. Unlike HFL that possess the original data and models locally, FTL makes the participants encrypt the gradients before data transmission, and adding random masks to avoid the participants from getting information of each other's. In healthcare applications, FTL can not only provide a standard framework, but can also achieve efficient classifications and avoid the problem of unbalanced classes \cite{130}.

\paragraph{Federated optimization} Compared with traditional machine learning, FL focuses on significant variability of system features on each device in the network (systems heterogeneity), and non-identically distributed data across the network (statistical heterogeneity) \cite{li2020federated}. Furthermore, aggregation FL requires defining an aggregation strategy, such as a method to combine the local models coming from the clients into a global one. For example, the standard and simplest aggregation strategy is federated averaging (known as FedAvg, \cite{mcmahan2017communication}). Another strategy is called FedProx \cite{stich2018local} is a generalization of FedAvg with some modifications to address data and the heterogeneity of the system. In addition, the FedProx is considered to explore the interactions between statistical and systematic heterogeneity \cite{li2020federated}. In addition, there is a new version of FedAvg called FedAvg+ based on personalization and cluster FL \cite{sattler2020clustered}. Also, FedEM model is based on the mixed distribution assumption with the expectation maximum algorithm \cite{marfoq2021federated}. Unfortunately, these optimization algorithms are still rarely applied in healthcare topics.

Table \ref{table3} reports FL techniques used recently for healthcare data. It observed that the  accuracy of FL model under the optimization algorithm is close to the central model, which demonstrates the potential of federated learning. However, the number of clients studied is still limited to 100 \cite{221}. Moreover, most of the data from these clients are derived from private datasets or multi-site datasets, which will limit the development of FL in practical medical applications \cite{hauschild2022federated}. So far, FL for healthcare is still in its early stages. The preliminary investigations provide a good foundation for future work, regardless of whether they involved replicating FL environments or doing limited experiments across hospitals \cite {chowdhury2022review}.

\subsection{Overview of methods and datasets used for XAI, DA, and FL}
Table \ref{table4} compares methods of XAI, DA and FL based on key strengths and weaknesses and lists code and datasets within each topic.
To overcome the limited availability of medical datasets, many of the XAI, DA and FL based on deep learning techniques employ transfer learning from a large dataset such as ImageNet \cite{5206848}.
Thus, we include the original datasets and code used for the recent and popular techniques.
For example, Grad-CAM is one of the common XAI methods used for multitask in medical topics like grading of ulcerative colitis \cite{158}, detecting cerebral hemorrhage \cite{101}, and predicting COVID-19 \cite{89,93} with a high degree of image interpretation. LIME also shows a feasible rate in multitask such as spinal posture classification \cite{136}, detection  of Parkinson's disease \cite{146}, and predicting depressive symptoms \cite{160}. Many other techniques used recently (e.g., Shapley Additive Explanations, Graph LIME, etc.) are almost based on deep learning models that need high power and performance computers.  

DA in high-dimensional complex data like medical imaging relies on deep learning to implicitly model nonlinear transformations\cite{14}.
For example, the difference of the feature distribution between source and target domains is narrowed by the deep domain confusion model for cross subject recognition \cite{zhang2019cross}.
Recent algorithms like Deep CORAL \cite{niu2021distant}, Deep Adaptation Networks \cite{chen2019synergistic}, Deep Subdomain Associate Adaptation Network (DSAAN) \cite{meng2022deep} and Emotional Domain Adversarial Neural Network (EDANN) \cite{tian2022novel} are recommended to reduce domain differences.
Despite the advances in DA models, few algorithms were applied to medical data\cite{14}.
An example in medical imaging is the explicit harmonization of uni-modal brain imaging data
across multiple sites with varying acquisition hardware\cite{https://doi.org/10.1002/jmri.27908      }.

Recent studies of FL provide promising results in healthcare applications.
For example, data from 20 institutes across the globe to train a model (EXAM: electronic medical record (EMR) chest X-ray AI model), for predicting COVID-19 \cite{213}. This model is based on HFL that considers differential privacy to mitigate risk of data ‘interception’ during site-server communication. In \cite{235}, three main models using standard transfer learning, FedAvg, and Federated Proximal frameworks, respectively, are considered to diagnose diabetic retinopathy. In \cite{bercea2022federated}, a new study proposes federated disentangled representation learning for unsupervised brain anomaly detection using MR scans from four different institutions. To solve the problems of security and trustworthiness, densely connected CNN based on FedAvg proposed and used for COVID-19 \cite{bai2021advancing}. Due to the limitations of sharing the data across hospitals and countries, more collaborations will improve the feasibility of FL in medical applications.

\begin{table*}[hbt!]
\setlength{\tabcolsep}{8pt}
\tiny
\caption{The strengthts and weaknesses of common XAI, DA, and FL techniques.}\label{table4}
\begin{tabular}{lm{4cm}m{4.2cm}ccc}

\hline
  \textbf{Technique} & \textbf{Strengths} & \textbf{Weaknesses} &  \textbf{Code} &  \textbf{Data} & \textbf{Reference}\\
\hline \hline 
\multicolumn{2}{c}{Explainable Artificial Intelligence} \\\hline \hline
\thead{Gradient-weighted \\Class Activation Mapping} &
\begin{itemize}[label=-]
    \item Simple to use 
    \item No need to change the network structure 
    \item Efficient to compute
    \item High degree of image explanation
\end{itemize} &
\begin{itemize}

\item The weight of saliency map depends on the
model and data set
\item Sometimes prominent areas cannot be correctly labeled
    \item Poor robustness
\end{itemize} &
 C1 & D1,D2 & \cite{Grad-CAM,158,101,89,93}  \\\hline

\thead{ Local Explainable \\ Model-agnostic Explanation} & 
\begin{itemize}
    \item Uses a simple model for local explanation
\end{itemize} & 
\begin{itemize}
    \item Poor robustness
    \item Inefficient calculation
\end{itemize} &
 C2& D3 & \cite{LIME,136,146,160}\\\hline

\thead{Graph LIME} &
\begin{itemize}
    \item Simple to use
    \item Trustworthy
\end{itemize} &
\begin{itemize}
    \item Inefficient calculation
    \item instable 
    \end{itemize} &
 *C3 & D4 & \cite{huang2022graphlime,133}\\\hline

\thead{Shapley Additive \\Explanations}  &
\begin{itemize}
    \item Useful for interpretation 
    \item Repeatability
\end{itemize} &
\begin{itemize}
    \item Inefficient computation
\end{itemize} &
 C4& D5,D6 & \cite{SHAP,143}\\\hline\hline

\multicolumn{2}{c}{Domain Adaptation} & \textbf{} & \textbf{} & \textbf{}\\\hline \hline
%\textbf{DA} & \textbf{} & \textbf{} & \textbf{} & \textbf{}\\\hline 

\thead{Deep Domain Confusion} & 
\begin{itemize} 
 \item Reduce the distribution difference between the source and target domains 
\end{itemize}&
  \begin{itemize} 
\item Only adapts to one layer of network
\item A single fixed kernel may not be the optimal kernel
\end{itemize}
&  C5 & D7 & \cite{tzeng2014deep,zhang2019cross}\\ \hline

\thead{Deep Adaptation Networks} & 
\begin{itemize} 
 \item Reduce domain differences 
\end{itemize}&
  \begin{itemize} 
\item Domain difference problem cannot be solved by semi-supervised learning
\end{itemize}
&  C5 & D7 & \cite{long2015learning,9514499}\\ \hline

%\cite{sun2016deep} & \thead{Deep CORAL} & 
%\begin{itemize} 
 %\item Nonlinear transformations
%\end{itemize}&
 % \begin{itemize} 
%\item It relies on linear transformations and is not end-to-end
%\end{itemize}
%&  C5 & D7\\ \hline

\thead{Deep Subdomain\\ Adaptation Network} & 
\begin{itemize} 
 \item The weight could be defined flexible
\end{itemize}&
  \begin{itemize} 
\item The weight value of network maybe could not be the best
\end{itemize}
&  C5 & D8 & \cite{zhu2020deep,jin2022multi} \\ \hline

\thead{Dynamic Adversarial\\ Adaptation Networks} & 
\begin{itemize} 
 \item Solve the dynamic distribution adaptation problem in adversarial networks
\end{itemize}&
  \begin{itemize} 
\item Can not handle regression problems
\end{itemize}
&  C5 & D7 & \cite{wang2020transfer,xu2022dynamic}\\ \hline\hline

\multicolumn{2}{c}{Federated Learning} & \textbf{} & \textbf{} & \textbf{}\\\hline \hline

\thead{EXAM / HFL } & 
\begin{itemize}
    \item Can use large, and heterogeneous data-sets
    \item Robust and generalizable
\end{itemize}&
\begin{itemize}
    \item Bias may occur due to limitations in data quality.
\end{itemize}&  C6 & D9 & \cite{213}\\\hline

\thead{AlexNet / FTL / \\ FedAvg \& FedProx} & 
\begin{itemize}
    \item Prediction rate is high
\end{itemize}& 
\begin{itemize}
    \item  Sub-optimal security
    \item  Data accuracy pitfalls
\end{itemize}&  - & D10-D12 & \cite{235}\\\hline

\thead{CNN / HFL / FedDis} &
\begin{itemize}
    \item Mitigate the statistical heterogeneity
    \item Good generalizability with clinical applications
     \item Flexibility 
\end{itemize}&
\begin{itemize}
       \item Privacy concerns
\end{itemize}& C7 & D13-D17 & \cite{bercea2022federated}\\\hline

\thead{3D-CNN / HFL / FedAvg} & 
\begin{itemize}
    \item High performance metrics
    \item Provided visual explanations
    %\item Analyzed the trade-offs between the model performance and the communication costs in the federated training process.
    \item Suitable for transfer learning
\end{itemize} &  
\begin{itemize}
    \item Weak federated training process for the unstable internet connection.
    \item Inefficient computation
\end{itemize} & C8 & D18 & \cite{bai2021advancing}\\ \hline

\end{tabular}

\footnotesize{FedDis: Federated Disentred representation learning for unsupervisived brain operatory segmentation; EXAM: Electronic Medical Record (EMR) chest X-ray AI model, "-” not available, “*” not official, C: code, D: data. 

C1  \url{https://github.com/zhoubolei/CAM},\url{https://github.com/Cloud-CV/Grad-CAM}\\
C2  \url{https://github.com/marcotcr/lime}\\
C3 \url{https://github.com/WilliamCCHuang/GraphLIME}\\
C4  \url{https://github.com/slundberg/shap}\\
C5  \url{https://github.com/jindongwang/transferlearning/ tree/master/code/DeepDA/} ,\url{https://github.com/ting2696/Deep-Symmetric-Adaptation-Network.}\\
C6 \url{https://ngc.nvidia.com/catalog/models/nvidia:med:clara_train_covid19_exam_ehr_xray }\\
C7  \url{https://github.com/albarqounilab/}\\
C8  \url{https://github.com/HUST-EIC-AI-LAB/UCADI}\\
D1  \url{http://host.robots.ox.ac.uk/pascal/VOC/voc2007}\\
D2 ImageNet Object Localization Challenge \cite{5206848}\\
D3 \url{https://github.com/marcotcr/lime-experiments}\\
D4 Datasets are from Cora and Pubmed.\\
D5 \cite{xiao2017fashion}\\
D6 \url{http://yann.lecun.com/exdb/mnist}\\
D7 \cite{saenko2010adapting,6975210,ZHUANG201677,KAVUR2021101950,yap2017automated}, \url{ultrasoundcases.info}\\
D8 \url{https://github.com/jindongwang/transferlearning/blob/master/data/dataset.md},  \cite{wang2020transfer}\\
D9 \url{https://www.kaggle.com/datasets/mariaherrerot/eyepacspreprocess}\\
D10 \url{https://www.adcis.net/en/third-party/messidor}\\
D11 \url{https://ieee-dataport.org/open-access/indian-diabetic-retinopathy-image-dataset-idrid}\\
D12 \url{https://www.kaggle.com/datasets/mariaherrerot/aptos2019}\\
D13 \url{https://www.oasis-brains.org}\\
D14 \url{http://adni.loni.usc.edu/data- 553 samples/access-data/}\\
D15 \url{http://lit.fe.uni-lj.si/tools.php?lang=eng}\\
D16 \url{https://smart-stats-tools.org/lesion-challenge- 556 2015}\\
D17 \url{https://www.med.upenn.edu/sbia/brats2018/data.html}\\
D18 \url{https://www.nhsx.nhs.uk/covid-19-response/data-and-covid-19/national-covid-19-chest-imaging-database-nccid/}\\}

\end{table*}

\section{Conclusion and future perspective} \label{S4}
The rate at which the complexity and volume of digital patient data is acquired, far exceeds the rate at which expert clinicians can be trained to analyze it.
Therefore, it is necessary to employ artificial intelligence to keep up with the growth of the data.
Deep learning based image analysis reached the performance of expert clinicians in making referral recommendations based on widespread available optical coherence tomography \mbox{\cite{Fauw2018}}. In many scenarios, however, artificial intelligence is not yet readily able to replace expert clinicians because the applicable AI methods are not plug-in replacements for human decision-making.
In neuroimaging, for example, the transition to the clinics did not take place despite many successes in research studies \mbox{\cite{Stephan.2017}}.
Application of AI with a human in the loop relieve the clinician from some--typically tedious, work. Atri-U, for example, is an application of AI in the clinical routine in which the AI back-end proposes a segmentation, two landmarks, and a time point to compute the volume of the left atrium \mbox{\cite{Anastasopoulos.2021}}. The clinician that bares the responsibility can either accept, modify, or ignore the proposals.
The latter scenario introduces a non-disruptive innovation into an existing diagnostic process with a substantially reduced average process time without compromising quality.
Further developments of such AI tools for clinical practice will depend on the key technologies discussed previously, and for each of which we will conclude with a summary.

\paragraph{Explainable AI} We found that the most XAI with deep learning addresses post-hoc interpretability to increase face value, like the CNN-CAM. We also found that the explainability levels are limited local explainability. This means that the model understanding can
only explain the model output rather than the internal decision-making.
Global explainability is desirable in clinical tasks to achieve trust.
More specifically, the practical XAI application is inevitable to consider people
who do not have the relevant AI background.
Additional explanations such as manual instructions will be greatly recommended.

\paragraph{Domain adaptation} Domain adaptation can effectively promote the transfer of models between different domains.
However, data heterogeneity and label validity are still a critical challenge. Many DA methods can eliminate the domain bias and the difference in data distribution, such as data-based methods and feature-based methods. Unfortunately, these methods are too rigid and require a lot of processing time. Unlike adversarial DA, which shows strong performance, it can effectively reduce domain differences and improve generalization performance. 

\paragraph{Federated learning} Federate learning is not well applied to healthcare data, and most studies focus on horizontal federated learning, which does not fully exploit the potential of federated learning. Only few optimization algorithms can achieve performance similar to the central model. We note that FL faces the challenges of poor data quality and insufficient model accuracy. This is due to the absence of standard and uniform data collection among institutions. Therefore, it mentions that the standard unified data with model security protection mechanism is the key to improving the deep learning model for clinical tasks.

\paragraph{Outlook to further challenges}
In the past decade, many methods originating in computer vision and other non-medical fields
found applications in medical research healthcare with some delay and minor adaptations.
If this trend continues, we expect to see more of the emerging XAI, DA, and FL methods in medicine
for tasks that are similarly structured as in non-medical fields.
We note, however, that healthcare processes often incorporate irregularly sampled, longitudinal, multi-modal data for decision-making.
The imputation of missing data can lead to inaccurate prediction that may result in biased estimations\cite{zhu2021predictive,pauzi2021comparison}.
With strong assumptions and simplified models, clinical decision processes can be modeled and optimized using mixed linear models that naturally handle missing data, as long as the input data is low-dimensional and in a homogeneous domain\cite{wyss2021adaptive}.
To train deep learning models with similar flexibility but more complex data,
much larger training data sets are required.
Combined, FA may contribute to leverage massive pools of private data that will be
transformed into a common reference domain with DA to train interpretable and thus more trustworthy
AI models using XAI.

\section*{Acknowledgment}
This work was supported in part by the National Natural Science Foundation of China (82260360) and the Foreign Young Talent Program (QN2021033002L).

% Bibliography and style
\bibliographystyle{unsrt}
\bibliography{reference}
%\bibliography{reference} 
%\bibliographystyle{IEEEtran}   
%\biboptions{numbers,sort&compress}

\end{document}